\def\paperID{2796}
\def\confName{CVPR}
\def\confYear{2025}

\def\paperTitle{\OurModel: Generalizable 3D-Language Feature Fields for Embodied Tasks}

\def\authorBlock{
    Zihan Wang \qquad
    Gim Hee Lee \\
    School of Computing, National University of Singapore \\
    {\tt zihan.wang@u.nus.edu}
}

\newif\ifreview 
\newif\ifarxiv \newcommand{\arxiv}{\arxivtrue}
\newif\ifcamera 
\newif\ifrebuttal 

\arxiv
\pdfoutput=1
\documentclass[10pt,twocolumn,letterpaper]{article}
\ifreview \usepackage[review]{cvpr} \fi
\ifarxiv \usepackage[pagenumbers]{cvpr} \fi
\ifrebuttal \usepackage[rebuttal]{cvpr} \fi
\ifcamera \usepackage{cvpr} \fi

\usepackage{graphicx}	
\usepackage{amsmath}	
\usepackage{amssymb}	
\usepackage{booktabs}
\usepackage{times}
\usepackage{microtype}
\usepackage{epsfig}
\usepackage[table,xcdraw,dvipsnames]{xcolor}
\usepackage{caption}
\usepackage{float}
\usepackage{placeins}
\usepackage{color, colortbl}
\usepackage{stfloats}
\usepackage{enumitem}
\usepackage{tabularx}
\usepackage{xstring}
\usepackage{multirow}
\usepackage{xspace}
\usepackage{url}
\usepackage{subcaption}
\usepackage{xcolor}
\usepackage[hang,flushmargin]{footmisc}
\usepackage{soul}

\ifcamera \usepackage[accsupp]{axessibility} \fi

% ...

\ifarxiv  \fi

\newcommand{\gh}[1]{{\color{black} {#1}}}
\newcommand{\OurModel}{{\color{black}{g3D-LF}}}

\newcommand{\R}[1]{{%
    \textbf{%
        \ifstrequal{#1}{1}{\textcolor{red}{R#1}}{%
        \ifstrequal{#1}{2}{\textcolor{blue}{R#1}}{%
        \ifstrequal{#1}{3}{\textcolor{magenta}{R#1}}{%
        \ifstrequal{#1}{4}{\textcolor{teal}{R#1}}{%
                           \textcolor{cyan}{R#1}%
        }}}}%
    }%
}}

\usepackage{xr-hyper}

\makeatletter
\newcommand*{\addFileDependency}[1]{
  \typeout{(#1)}
  \@addtofilelist{#1}
  \IfFileExists{#1}{}{\typeout{No file #1.}}
}

\makeatother
\newcommand*{\myexternaldocument}[1]{
    \externaldocument{#1}
    \addFileDependency{#1.tex}
    \addFileDependency{#1.aux}
}

\definecolor{cvprblue}{rgb}{0.21,0.49,0.74}
\usepackage[pagebackref,breaklinks,colorlinks,citecolor=cvprblue]{hyperref}
\usepackage[capitalize]{cleveref}
\crefname{section}{Sec.}{Secs.}
\crefname{table}{Table}{Tables}
\crefname{figure}{Fig.}{Figs.}

\ifarxiv \crefname{appendix}{App.}{Apps.}
\else \crefname{appendix}{Suppl.}{Suppls.} \fi

\usepackage{pifont}
\usepackage{hyperref}
\unless\ifarxiv \myexternaldocument{_supplementary} \fi

\begin{document}
\title{\paperTitle}
\author{\authorBlock}
\maketitle

\begin{abstract}
\gh{We introduce} Generalizable 3D-Language Feature Fields (
\OurModel{}), a 3D representation model pre-trained on large-scale 3D-language dataset for embodied tasks. 
\gh{Our \OurModel{}} 
processes posed RGB-D images from agents to encode feature fields 
\gh{for: 1) Novel view representation predictions from any position in the 3D scene; 2) Generations of BEV maps centered on the agent; 3) Querying targets using multi-granularity language within the above-mentioned representations.} 
\gh{Our representation} can be generalized to unseen environments, enabling real-time construction and dynamic updates. By volume rendering latent features along sampled rays and integrating semantic and spatial relationships through multiscale encoders, 
\gh{our \OurModel{}} 
produces representations at different scales and perspectives, aligned with multi-granularity language, via multi-level contrastive learning. \gh{Furthermore, we prepare a large-scale 3D-language dataset to align the representations of the feature fields with language.} Extensive experiments on Vision-and-Language Navigation \gh{under} both Panorama and Monocular settings, Zero-shot Object Navigation, and Situated Question Answering tasks highlight the significant advantages and effectiveness of our 
\OurModel{} 
for embodied tasks. The code is available at
\href{https://github.com/MrZihan/g3D-LF}{https://github.com/MrZihan/g3D-LF}.
\end{abstract}
\section{Introduction}
\label{sec:intro}

Embodied agents seek to understand 3D environments, enabling interaction with environments and human 
\gh{by performing} tasks such as Question Answering~\cite{masqa3d,azuma2022scanqa,majumdar2024openeqa}, Navigation~\cite{chaplot2020object,majumdar2022zson,yokoyama2024vlfm,anderson2018vision,krantz2020beyond,kwon2023renderable}, \etc.
To this end, various 3D scene representation models tailored for embodied tasks have been proposed, including point cloud-based models~\cite{zhu20233d,huang2023embodied,chen2024grounded}, 3D occupancy~\cite{liu2024volumetric}, hybrid voxel~\cite{fu2024scene}, and feature fields~\cite{shen23Distilled,ze2023gnfactor,wang2024lookahead,qiu2024learning}.

\begin{figure}
\noindent\begin{minipage}[h]{1\columnwidth}%
\begin{center}
\includegraphics[width=1.1\columnwidth]{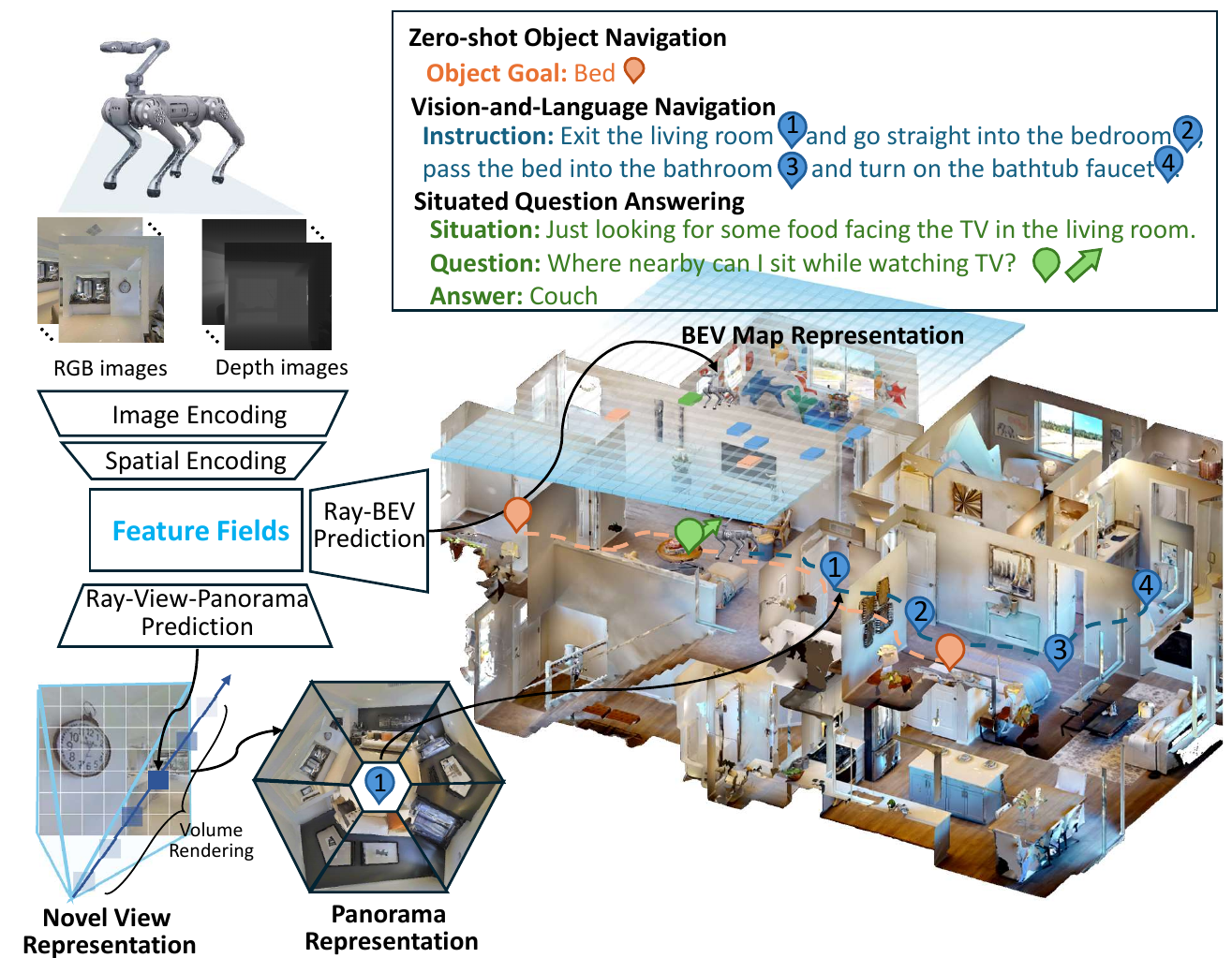}
\par\end{center}%
\end{minipage}
\caption{
\gh{Our \OurModel{} uses posed RGB-D images from the agent to predict novel view and BEV map representations at various scales within the 3D scene, aligned with multi-granularity language through 3D-language pre-training. The representation is
}
applicable to embodied tasks like visual navigation and embodied question answering, facilitating scene representation, language-guided querying, and navigation planning.}
\label{fig:introduction}
\vspace{-6mm}
\end{figure}

For multimodal embodied tasks in large-scale scenes, 3D representation models typically need: 1) generalization to unseen scenes, 2) construct and update representations in real time, and 3) open-vocabulary semantic space. The generalizable 3D feature fields provides the above advantages and has been widely explored across various embodied tasks. Unlike point cloud-based models that depend on complete and low-noise point clouds 
which are 
less robust, the implicit representations of the feature fields are derived from the 2D foundation model, preserving semantic expressiveness even with few-shot observations from 3D scenes. As shown in Figure~\ref{fig:introduction}, the feature fields model uses RGB-D images as input to encode and update implicit scene representations, which are then used to predict novel view, panorama and BEV map representations associated with language through volume rendering. 
\gh{These predicted representations can assist embodied tasks such as navigation planning~\cite{wang2024lookahead,wang2024simtoreal,qiu2024learning}, \etc.}
%
\gh{However, several significant drawbacks remain in these feature fields models:}
1) The supervision for the predicted representations comes from 2D foundation models, \textit{e.g.}, CLIP~\cite{radford2021learning} and DINOv2~\cite{oquab2024dinov2} 
greatly limits the understanding for 3D spatial relationships; 2) These models are trained without language supervision, resulting in a substantial gap with language semantics; 3) The large-scale representations, \textit{e.g.}, panorama and BEV map from feature fields is particularly challenging for long text understanding. These issues severely limit the potential of the feature fields model on language-guided embodied tasks.

\gh{To circumvent the above-mentioned issues, we introduce Generalizable 3D-Language Feature Fields (\OurModel{}), a 3D representation model pre-trained on large-scale 3D-language dataset for embodied tasks.}
We \gh{first curate and consolidate} 
a large amount of 3D-language data from previous works~\cite{jia2024sceneverse,chen2020scanrefer,zhang2024vla3ddataset3dsemantic} to train our \OurModel{} model. 
\gh{These data include} 5K indoor scenes and almost 1M language descriptions of multiple granularities. The text annotations include object categories, object characteristics, object relationships, and the spatial layout of the entire scene, which are employed to supervise multiscale encoders of the \OurModel{} model. 
\gh{We then design our \OurModel{} model to learn generalizable 3D-language feature fields.}
\gh{To this end, we employ multi-level contrastive learning for multi-scale encoders to align predicted representations and language across different scales.}
For the regional representation within the novel view, 
\gh{a} contrastive loss is calculated across 1,883 indoor object categories. For the predicted novel view representation, both the CLIP visual representations and language are employed for contrastive training to balance generalization ability and language alignment. For large-scale panorama and BEV representations, we propose the fine-grained contrastive learning based on the affinity matrix to achieve long text understanding.

The pre-trained \OurModel{} model is 
\gh{subsequently} evaluated on various embodied tasks, including vision-and-language navigation (monocular setting~\cite{wang2024simtoreal} and panorama setting~\cite{wang2024lookahead}), zero-shot object navigation~\cite{yokoyama2024vlfm}, and situated question answering~\cite{masqa3d}, gains significant performance improvements.
In this work, our \textbf{main contributions} include:

\begin{itemize}
\item We organize a large-scale 3D-language dataset to train the feature fields model.

\item This work proposes the Generalizable 3D-Language Feature Fields (\OurModel{}) \gh{with} 
a multi-level contrastive learning framework to align the multi-scale representations of feature fields with multi-granularity language.

\item 
\gh{Our proposed \OurModel{}} model improves multiple baseline methods to state-of-the-art performance across various embodied tasks, 
\gh{thus validating} the potential of \gh{our} generalizable feature fields for Embodied AI.
\end{itemize}


\section{Related Work}
\label{sec:related}

\noindent \textbf{Generalizable 3D Feature Fields.} 
The neural radiance field (NeRF)~\cite{mildenhall2021nerf} has gained significant popularity in various AI tasks, which predicts the RGB image from an arbitrary viewpoint in a 3D scene. Furthermore, some works leverage NeRF-based methods to predict novel view representations instead of RGB values, enabling 3D semantic segmentation~\cite{voranesf} and 3D language grounding~\cite{kerr2023lerf}. 
However, these methods with implicit MLP networks can only synthesize novel view representations in seen scenes, which makes it difficult to generalize to unseen large-scale scenes and adapt to many embodied AI tasks (\textit{e.g.}, navigation). To this end, some works~\cite{wang2024lookahead,qiu2024learning,taioli2023language} attempt to encode 2D visual observations into 3D representations (called Generalizable 3D Feature Fields) via the depth map. Through volume rendering~\cite{mildenhall2021nerf}, these models decode novel view representations from the feature fields and align them with open-world features (\textit{e.g.}, CLIP embeddings~\cite{radford2021learning}). The 3D feature fields can generalize to unseen scenes, enabling real-time construction and dynamic updates. However, the drawback of these models lies in the fact that the supervision of their predicted representations comes from 2D visual models, which limits their performance in language-guided embodied tasks. Our work offers a feasible approach to training the 3D feature fields model with large-scale 3D-language data.

\vspace{2mm}
\noindent \textbf{Vision-and-Language Navigation.} 
Vision-and-Language Navigation (VLN)~\cite{anderson2018vision,krantz2020beyond,zhang2024vision,hong2021vln,chen2021history,qiao2023hop+,wang2024vision} requires the agent understand complex natural language instructions and navigate to the described destination using low-level actions, \textit{e.g.}, turn left 15 degrees, turn right 15 degrees, or move forward 0.25 meters. To address inefficiencies and poor performance in atomic action prediction, some works~\cite{krantz2022sim2sim,Hong2022bridging,wang2024simtoreal} develop waypoint predictors to generate several candidate waypoints around the agent. The navigation policy model can then select the optimal waypoint as the next sub-goal and execute atomic actions to move, greatly enhancing planning efficiency. In this context, how to represent waypoints and carry out planning have become critical. Some works use a topological map~\cite{chen2022think,an2024etpnav} or BEV map~\cite{an2023bevbert,wang2023gridmm,liu2023bird} to represent semantic relationships between waypoints, while some~\cite{wang2024lookahead,wang2024simtoreal} explore feature fields to predict waypoint representations of novel views and improve navigation planning. 
\gh{Our \OurModel{} model} further improves the performance of methods using feature fields.

\vspace{2mm}
\noindent \textbf{Zero-shot Object Navigation.}
In object-goal navigation~\cite{chaplot2020object,ramakrishnan2022poni,zhang2021hierarchical}, an agent is tasked with locating a specified object within indoor environments. Typically, reinforcement learning~\cite{zhu2017target} is used to train a policy network that predicts actions, while object detection~\cite{wang2023yolov7,liu2023grounding} or segmentation models~\cite{kirillov2023segment,zhang2023faster,he2017mask} help identify the object. However, these navigation models are often limited to specific objects, making open-vocabulary navigation challenging and hindering generalization in real-world applications~\cite{gervet2023navigating}. To address this \gh{issue}, zero-shot navigation methods have emerged~\cite{majumdar2022zson,zhou2023esc,gadre2023cows,yokoyama2024vlfm}, leveraging Vision-and-Language Models (VLMs)~\cite{radford2021learning,li2022grounded,li2023blip} to identify potential directions or areas containing the target, 
\gh{followed by using} the pre-trained pointgoal navigation models~\cite{wijmans2019dd} to search the potential areas. Considering that general 2D VLMs are not fully suited for indoor 3D environments \gh{and to the best of our knowledge}, we are the first to attempt using the indoor 3D feature fields model for zero-shot object navigation.

\vspace{2mm}
\noindent \textbf{Situated Question Answering.} The Embodied Question Answering tasks~\cite{azuma2022scanqa,das2018embodied,majumdar2024openeqa} require the agent to observe the 3D environment and answer questions from humans. Furthermore, Situated Question Answering~\cite{masqa3d} requires \gh{advanced 3D spatial understanding of} the agent 
to answer the question 
\gh{and to} interpret and locate the position and orientation of the textual description.
Compared to previous works~\cite{huang2023embodied,jia2024sceneverse,fu2024scene} using point clouds, we only use RGB-D images to encode feature fields and leverage their multi-scale representations for localization and question answering.

\section{\gh{Our} Method}
\label{sec:method}

\subsection{3D-Language Data}
\gh{We prepare a large-scale 3D-language dataset to align the representations of the feature fields with language.}
Our dataset includes about 5K 3D indoor scenes, mainly sourced from the single-room scans ScanNet~\cite{dai2017scannet}, multi-room house scans of the Habitat-Matterport 3D dataset (HM3D)~\cite{ramakrishnan2habitat,yadav2023habitat}, and the photo-realistic multi-room scenes of Structured3D~\cite{Structured3D}. The total number of language annotations is close to one million, which are mainly sourced from the SceneVerse dataset~\cite{jia2024sceneverse}. SceneVerse uses 3D scene graphs and large language models (LLMs) to automate high-quality object-level and scene-level descriptions. 
The annotations also includes the large set of human-annotated object referrals~\cite{chen2020scanrefer}.

\gh{We organize the dataset as follows to streamline feature fields training:}
1) For each 3D scene, the agent can observe numerous RGB-D images 
\gh{and its} corresponding poses as inputs. 
2) An instance-level point clouds mark each instance in the scene with an instance ID 
which can be used to retrieve associated language descriptions from the database.
It is \gh{thus} easy to 
\gh{get instances that} are near any given point in the 3D scene and obtain their language descriptions. This enables \gh{the} training code to efficiently obtain language annotations for specific regions within a novel view or a BEV map.

\begin{figure*}[ht]
\makebox[\textwidth][c]
{\includegraphics[width=0.8\paperwidth]{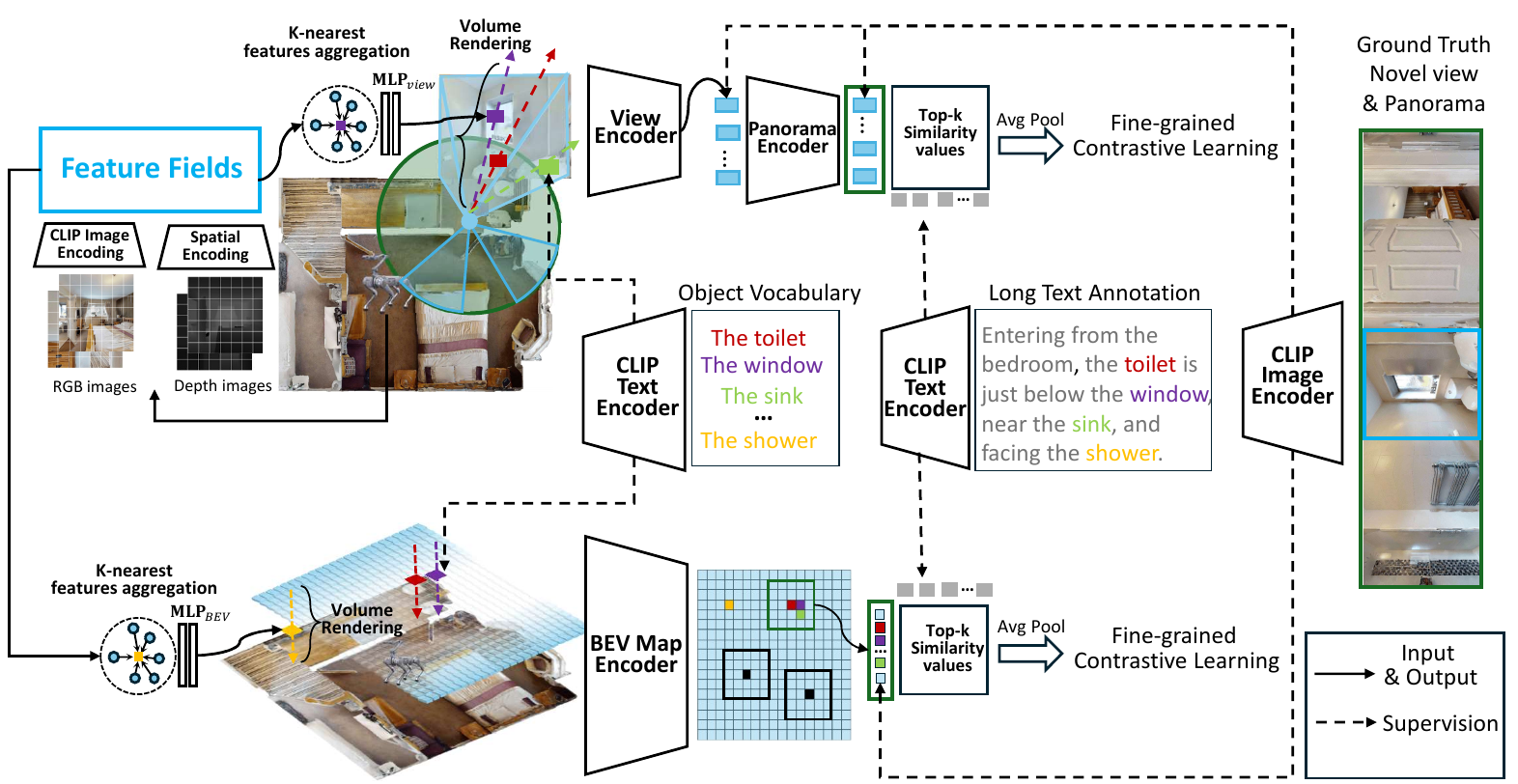}}
\vspace{-20pt}
\caption{
\textbf{\gh{Overview of our \OurModel{} model.}} 
\gh{Our} 
model encodes the observed RGB-D images into the feature fields (consists of many feature points). Through aggregating k-nearest features, the MLP networks predict the latent feature and volume density of sampled points along the rendered ray. The hierarchical encoders further generate representations of novel view, panorama, and BEV map, then conduct multi-level contrastive learning with multi-granularity language.}
\label{fig:framework}
\vspace{-10pt}
\end{figure*}

\subsection{3D-Language Feature Fields} 
\noindent \textbf{Feature Fields Encoding.}
\gh{As shown in Figure~\ref{fig:framework}, our \OurModel{} model follows HNR~\cite{wang2024lookahead} to take a posed RGB image as input and uses the CLIP image encoder to extract fine-grained visual features $\{\textbf{g}_{t,i}\in {\mathbb{R}}^{768}\}_{i=1}^I$. $\textbf{g}_{t,i}$ denotes the $i$-th feature patch of the CLIP feature map extracted from $t$-th frame observed by the agent.
We then map $\textbf{g}_{t,i}$ to the corresponding 3D world coordinates $\{P_{t,i}\}_{i=1}^I$ using the depth map and camera parameters.}

For each feature $\textbf{g}_{t,i}$, the observed horizontal orientation $\theta_{t,i}$ and the regional size $s_{t,j}$ are also calculated and stored \gh{to enhance the spatial representation}. 
The set of feature points $\mathcal{M}$ can \gh{therefore} be dynamically updated \gh{as}:
%
\begin{equation}
    \mathcal{M}_{t} = \mathcal{M}_{t-1} \cup \{[\textbf{g}_{t,i}, P_{t,i}, \theta_{t,i}, s_{t,i}]\}_{i=1}^{I} \text{.}
\end{equation}
\noindent \textbf{Ray-View-Panorama Encoding.}
The $\textbf{MLP}_{view}$ network 
aggregates nearby features within feature fields $\mathcal{M}$ and encode their spatial information~\cite{wang2024lookahead} (
\ie, relative positions and relative directions) \gh{to predict semantic representations $\textbf{r}\in {\mathbb{R}}^{768}$ and volume density $\sigma\in {\mathbb{R}}^{1}$ at any point from any direction in the continuous fields.}

For each novel view, 
\gh{our} \OurModel{} model generates a feature map $\textbf{R}\in {\mathbb{R}}^{12\times12\times768}$ by predicting subregion features through volume rendering within feature fields. The model samples \(N\) points along the ray from the camera position to each subregion 
center 
\gh{to search for} the k-nearest features and predicting volume density \(\sigma_n\) and latent representation \(\textbf{r}_n\), which then are composited into a subregion feature: 
%
%
\begin{equation}
    \begin{split}
        \textbf{R}_{(u,v)} &= \sum_{n=1}^{N}\tau_n(1-\exp(-\sigma_n\Delta_n))\textbf{r}_n, \\ 
        \text{where} \qquad \qquad
        \tau_n &= \exp({-\sum_{i=1}^{n-1}\sigma_i\Delta_i}).
    \end{split}
    \label{equ:volume_rendering}
\end{equation}
Here, \(\tau_n\) represents volume transmittance 
and \(\Delta_n\) is the distance between sampled points. \(\textbf{R}_{(u,v)}\) denotes the regional feature at the \( u \)-th row and \( v \)-th column of the novel view feature map \(\textbf{R}\).
%
\gh{We integrate context of the surrounding by feeding the feature map $\textbf{R}$ together with a learnable view token $\textbf{V}\in {\mathbb{R}}^{768}$ into the transformer-based view encoder to obtain the encoded $\textbf{R}'$ and novel view representation $\textbf{V}'$ that represent the entire novel view.}
Furthermore, to reason relationships across multiple views within a panorama, 
\gh{our \OurModel{}} model predicts 12 novel views $\{\textbf{V}_{i}'\}_{i=1}^{12}$ around the viewpoint at 30-degree intervals and combines them into a transformer-based panorama encoder to obtain $\{\textbf{V}_{i}''\}_{i=1}^{12}$.


\vspace{2mm}
\noindent \textbf{Ray-BEV Encoding.}
\gh{The novel view and panorama representations are insufficient for larger-scale scene understanding. To circumvent this problem, we propose to construct BEV map representation via our \OurModel{} as shown in Figure~\ref{fig:framework}.}
Unlike novel view prediction 
where rays are emitted from the viewpoint along the viewing cone, the rendering rays for the BEV map are rendered vertically from top to bottom. 
\gh{The starting point of the rendered ray is set slightly below the ceiling to avoid being blocked.}

Specifically, the $\textbf{MLP}_{BEV}$ network is used to aggregate the nearest feature points to the sampled point and predict its semantic representation $\hat{\textbf{r}}_n$ and volume density $\hat{\sigma}_n$ in the continuous field. Subsequently, the ray representation $\hat{\textbf{R}}_{(h,w)}\in {\mathbb{R}}^{768}$ can be obtained using the similar volume rendering method of Equation~\ref{equ:volume_rendering}, where $(h,w)$ denotes the $h$-th row and $w$-th column of the BEV map $\hat{\textbf{R}}\in {\mathbb{R}}^{168\times168\times768}$. To cover the large scene, the BEV map $\hat{\textbf{R}}$ encompasses a \( 16.8 \, \text{m} \times 16.8 \, \text{m} \) area centered on the agent.
After downsampling the BEV map to $\hat{\textbf{R}}_{conv}\in {\mathbb{R}}^{24\times24\times768}$ through a non-overlapping $7\times7$ convolution layer, the transformer-based BEV map encoder captures semantic relationships between different regions 
\gh{to get} the encoded BEV map representations $\hat{\textbf{R}}'\in {\mathbb{R}}^{24\times24\times768}$.

\subsection{Multi-level Contrastive Learning}\label{sec:multi_CL}

\noindent \textbf{Balanced Object-level  Alignment.}
We apply contrastive supervision using an object vocabulary $\mathcal{O} \in \mathbb{R}^{1883 \times 768}$ that spans 1,883 indoor object categories
\gh{for supervision of the $\textbf{MLP}_{view}$ and $\textbf{MLP}_{BEV}$ networks to predict latent features in feature fields.} For ray representations $\textbf{R}$ obtained via volume rendering, the cosine similarities $\{\text{CosSim}(\textbf{R}, \mathcal{O}_i)\}_{i=1}^{1883}$ are computed with each vocabulary embedding. The training objective is to 
\gh{maximize and minimize similarity for the correct and other object category, respectively, \ie:}
\begin{equation}
    \mathcal{L}_{object} = \operatorname{CrossEntropy}(\{
    \operatorname{CosSim}(\textbf{R},\mathcal{O}_i)/\tau\}_{i=1}^{1883}, \mathcal{O}^{gt}),
\end{equation}
\gh{where} \( \mathcal{O}^{gt} \) denotes the ground-truth category and \( \tau \) is the temperature coefficient for contrastive learning. Similarly, the object alignment loss for the ray representations $\hat{\textbf{R}}$ of the BEV map 
denoted as \( \hat{\mathcal{L}}_{object} \) 
can also be calculated.

\gh{We notice the network struggles to recognize smaller objects such as the \textit{lamp} due to the dominance of some objects (\textit{e.g.}, \textit{floor} and \textit{walls}) leading to long-tailed distribution in the indoor scenes.}
To address this \gh{issue}, we implement a balanced loss that emphasizes harder-to-recognize objects. Specifically, 
the weight of loss for the rays of top 10\% cross entropy 
\gh{are} significantly increased using a scaling factor \( \alpha \) \gh{for ray representations within the novel view or BEV map}. In short, rays with higher cross entropy indicate harder-to-recognize objects and therefore have a higher loss weight.

\vspace{2mm}
\noindent \textbf{Fine-grained Contrastive for Long Text.}
To enable 
\gh{our} \OurModel{} model to understand object relationships and spatial layouts, we propose a fine-grained contrastive learning method for long text alignment. As shown in Figure~\ref{fig:framework}, \gh{our \OurModel{}} aligns the BEV features in a window (\textit{e.g.}, $5\times5$) with the long text features 
\gh{to enhance} the representation of the BEV map for spatial semantics. 
Specifically, centered on an instance, the BEV features $\{\hat{\textbf{R}}'_{i}\}_{i=1}^{25}$ within the window are associated with $L$ word features $\{\textbf{W}_{l}\}_{l=1}^{L}$ from the CLIP text encoder through an affinity matrix $\mathbf{A}$:
\begin{equation}
    \mathbf{A}_{(i,l)} = \text{CosSim}(\hat{\textbf{R}}'_{i},\textbf{W}_{l})/\tau.
\end{equation}
%
The highest $L$ similarity scores (equal to the number of words) are extracted from the affinity matrix $\mathbf{A}$, and their average is used as the fine-grained similarity score between the BEV window and the long text features:
\begin{equation}
    \operatorname{FineSim}(\{\hat{\textbf{R}}'_{i}\}_{i=1}^{25},\{\textbf{W}_{l}\}_{l=1}^{L})=\operatorname{Avg}(\operatorname{Topk}(\mathbf{A},L)).
\end{equation}
%
\gh{Denoting the} BEV features within the $i$-th window as $\mathbf{B}_i$ and the $j$-th text features as $\mathbf{T}_j$, the fine-grained contrastive learning loss can be calculated as:
\begin{equation}
\small
    \begin{split}
        \hat{\mathcal{L}}_{long\_text}&=\frac{1}{J} \sum_{j=1}^{J}\operatorname{CrossEntropy}(\{\operatorname{FineSim}(\mathbf{B}_i,\mathbf{T}_j)\}_{i=1}^{I}, j) \\
        &+ \ \frac{1}{I} \sum_{i=1}^{I}\operatorname{CrossEntropy}(\{\operatorname{FineSim}(\mathbf{T}_j, \mathbf{B}_i)\}_{j=1}^{J},i).
    \end{split}
\end{equation}
Similarly, 
\gh{our} \OurModel{} model performs fine-grained contrastive learning between encoded panoramic representations $\{\textbf{V}_{i}^{''}\}_{i=1}^{12}$ and long-text features $\{\textbf{W}_{l}\}_{l=1}^{L}$  to compute the fine-grained contrastive loss $\mathcal{L}_{long\_text}$.

\vspace{2mm}
\noindent \textbf{CLIP Knowledge Distillation.}
Since the 3D-language data is orders of magnitude smaller than image-language data (millions vs. billions~\cite{radford2021learning}), 
\gh{our} \OurModel{} model still distills visual features from CLIP model~\cite{radford2021learning} to ensure robust generalization. Specifically, 
\gh{our} \OurModel{} uses CLIP features extracted from the ground-truth novel view or corresponding region image for contrastive supervision \gh{on the predicted new view representation $\textbf{V}^{'}$, the panorama representation $\textbf{V}_{i}^{''}$, and the BEV map representation $\hat{\textbf{R}_{i}}'$, \ie}:
\begin{equation}
\small
  \mathcal{L}_{view\_clip}=\frac{1}{I} \sum_{i=1}^{I}\operatorname{CrossEntropy}(\{{\operatorname{CosSim}}(\textbf{V}_{i}',\textbf{V}_{j}^{gt})/\tau\}_{j=1}^{J}, i),  
\end{equation}
where \(\textbf{V}_{j}^{gt}\) denotes the ground truth CLIP feature for $j$-th novel view representation \(\textbf{V}_{j}'\). Similarly, the contrastive loss $\mathcal{L}_{pano\_clip}$ for \gh{the} panoramic representation and $\mathcal{L}_{bev\_clip}$ for the BEV map can also be computed. 

\subsection{Embodied Tasks}
To verify the effectiveness of 
\gh{our} \OurModel{} model for embodied tasks, 
\gh{we integrate} the predicted representations from 
\gh{our model} into existing baseline methods and evaluates performance on Vision-and-Language Navigation, Zero-shot Object Navigation, and Situated Question Answering tasks.

\begin{figure}
\noindent\begin{minipage}[h]{1\columnwidth}%
\begin{center}
\includegraphics[width=1.0\columnwidth]{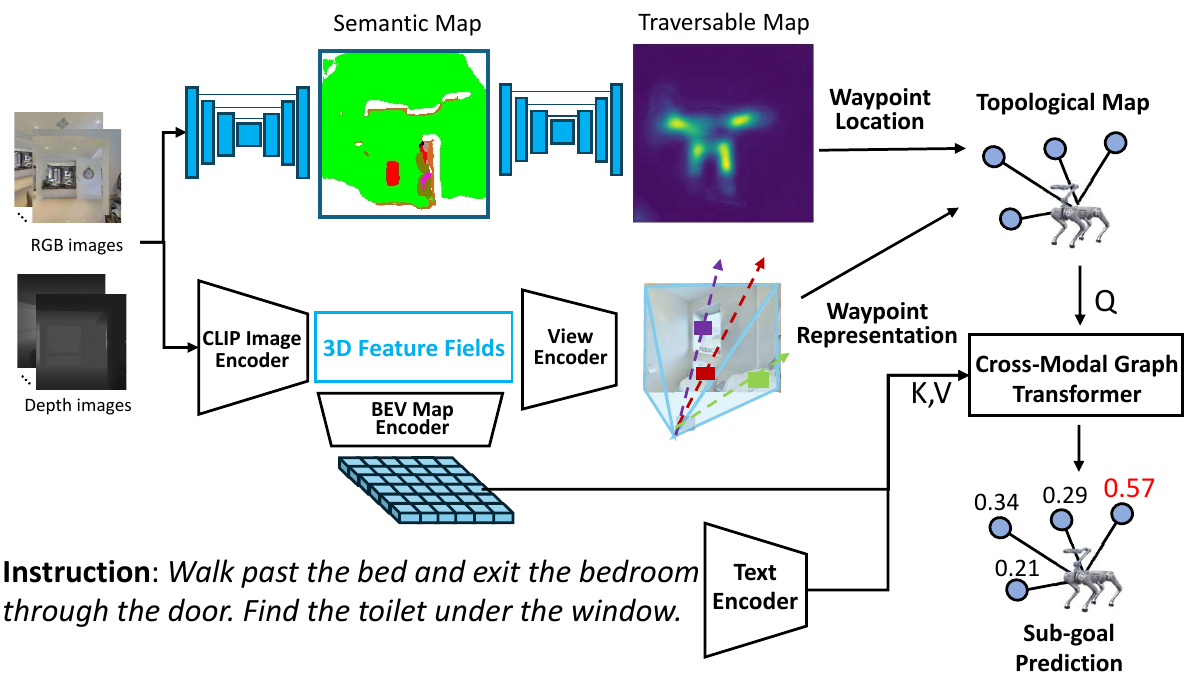}
\par\end{center}%
\end{minipage}
\vspace{-3mm}
\caption{
\gh{Monocular VLN framework}
based on VLN-3DFF~\cite{wang2024simtoreal}.}
\label{fig:vln}
\vspace{-5pt}
\end{figure}

\vspace{2mm}
\noindent \textbf{Vision-and-Language Navigation.} We evaluate the \OurModel{} model on VLN tasks with two settings. The first setting is with the monocular camera, which only allows the agent to observe the forward-facing view. As shown in Figure~\ref{fig:vln}, the VLN-3DFF~\cite{wang2024simtoreal} is a monocular VLN model that predicts candidate waypoints around the agent using a semantic map, and predicts each candidate's representation with generalizable feature fields~\cite{wang2024lookahead} and then selects the optimal waypoint to move through a cross-modal graph encoder~\cite{chen2022think,an2024etpnav}. Based on this baseline method, we incorporate novel view representations from our \OurModel{} model and input the BEV map into the cross-modal graph encoder following GridMM~\cite{wang2023gridmm} to enhance spatial layout understanding.
The second setting is with the panorama camera, in which the agent can observe 12 RGB-D view images within the panorama. Following HNR~\cite{wang2024lookahead},
a waypoint predictor~\cite{Hong2022bridging} is used to predict candidate waypoints, and our \OurModel{} model generates panorama representations of these waypoints for navigation planning.

\begin{figure}
\noindent\begin{minipage}[h]{1\columnwidth}%
\begin{center}
\includegraphics[width=1.0\columnwidth]{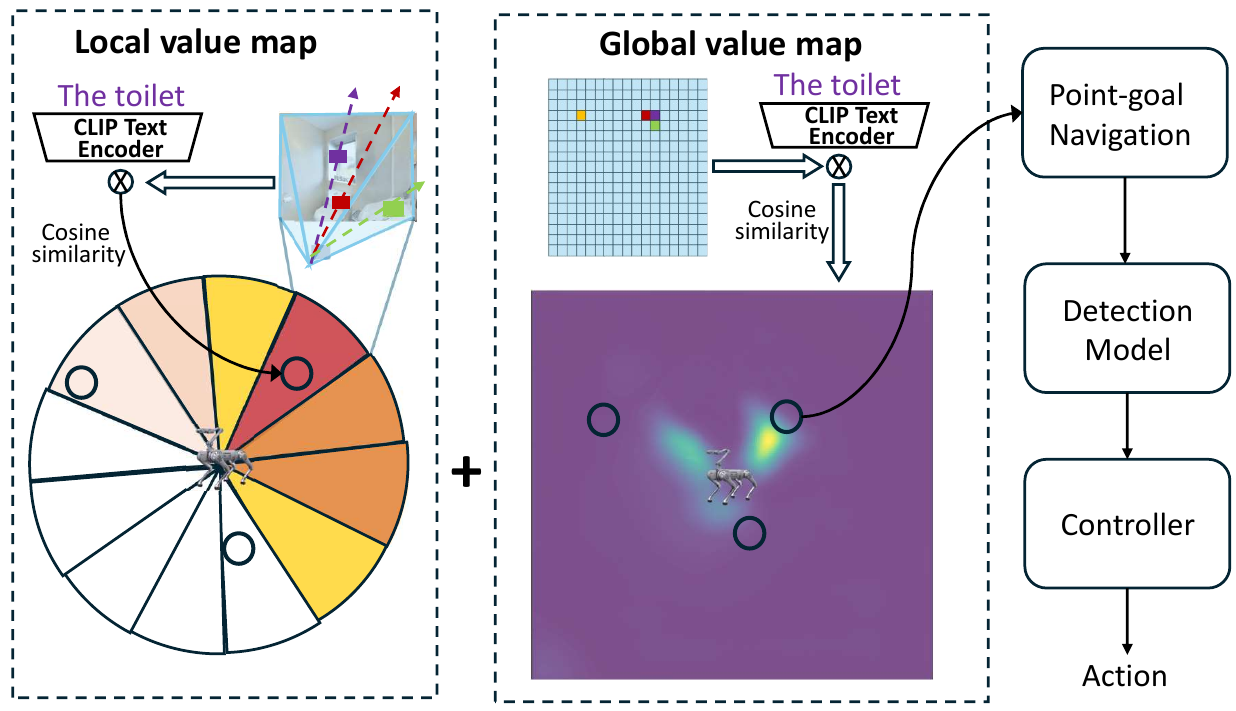}
\par\end{center}%
\end{minipage}
\vspace{-3mm}
\caption{
\gh{Zero-shot object navigation framework} based on VLFM~\cite{yokoyama2024vlfm}.}
\label{fig:zson}
\vspace{-5mm}
\end{figure}

\vspace{2mm}
\noindent \textbf{Zero-shot Object Navigation.}
As shown in Figure~\ref{fig:zson}, unlike the baseline method VLFM~\cite{yokoyama2024vlfm} that uses the 2D foundation model BLIP-2~\cite{li2023blip} to calculate the similarity between the target object and visual observations to construct the value map, we use \gh{our} \OurModel{} to predict the value of potential regions. Although the monocular agent can only observe the forward view, our \OurModel{} predicts 12 novel view feature maps surrounding the agent within panorama based on historical observations, and calculates max similarity in feature map with the target object. 
The text features of the target object are also used to calculate the similarity with each region representation on the BEV map \gh{to obtain a larger-scale value map}. Combining these two value maps, the navigation agent prioritizes traveling to the candidate waypoint with the highest similarity score.

\begin{figure}
\noindent\begin{minipage}[h]{1\columnwidth}%
\begin{center}
\includegraphics[width=1.0\columnwidth]{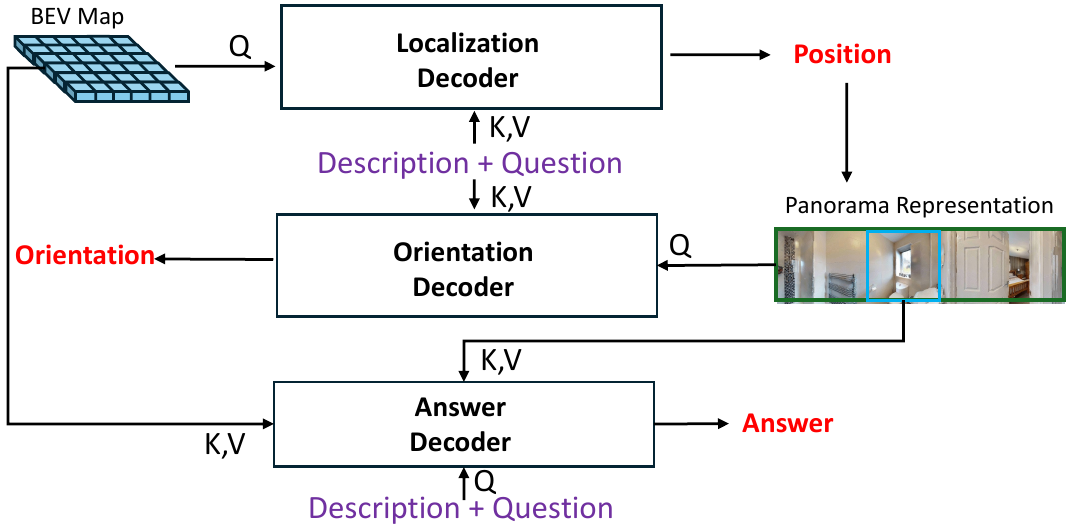}
\par\end{center}%
\end{minipage}
\caption{The framework of 
situated question answering~\cite{masqa3d}.}
\label{fig:sqa3d}
\vspace{-5mm}
\end{figure}

\vspace{2mm}
\noindent \textbf{Situated Question Answering.}
\gh{A three-stage framework is shown in Figure~\ref{fig:sqa3d}, where we use our \OurModel{} to train three transformer-based decoders for position, orientation and answer predictions.} 
First, the Localization Decoder predicts the heatmap for location of the textual description based on the BEV map. 
Our \OurModel{} model generates the panorama representations around 
\gh{the predicted location}, which are then processed by the Orientation Decoder to predict the orientation. Finally, the textual description, question, BEV map, and panorama representations are fed into the Answer Decoder to generate the final answer.

\section{Experiments}
\label{sec:experiments}

\subsection{Experiment Setup and Metrics}
\noindent \textbf{
\OurModel{} Pre-training.} 
\gh{We pre-train our \OurModel{} model shown in Figure~\ref{fig:framework} on 5K 3D scenes.} During training, 
30 frames are uniformly sampled from the RGB-D video \gh{of each scene in the ScanNet~\cite{dai2017scannet} dataset} to construct the feature fields, with an additional frame randomly selected as the novel view for prediction. The \OurModel{} then predicts the panorama representation and BEV map centered on the camera of this novel view. For each ray in the novel view or BEV map, the corresponding instance ID can be searched by calculating the nearest instance point to the rendered surface within the annotated instance point cloud. 
The language annotations of the novel view, panorama, and BEV map can \gh{thus} be obtained by retrieving language annotations with their instance IDs from the database for training. 
\gh{Due to the limited number of images per scene (fewer than 20), we use all available images from the Structured3D~\cite{Structured3D} dataset for training.}
\gh{We follow HNR~\cite{wang2024lookahead} for the HM3D~\cite{ramakrishnan2habitat,yadav2023habitat} dataset using the Habitat simulator~\cite{savva2019habitat} to randomly sample navigation trajectories and the observed RGB-D images to predict the novel views and panoramas around candidate waypoints, and construct the BEV map centered on the agent.}
The multi-level contrastive losses described in Section~\ref{sec:multi_CL} are utilized to optimize the 
\OurModel{} model.

\gh{Finally,} we combine scenes from all datasets and pretrain 
\gh{our} \OurModel{} model for 50K episodes (about 10 days) on two RTX 6000 Ada GPUs. \textbf{To ensure fair comparisons on downstream tasks, all training data only includes the train split, the val and test splits are removed}.

\vspace{2mm}
\noindent \textbf{Vision-and-Language Navigation.} We evaluate the VLN model on the VLN-CE dataset~\cite{krantz2020beyond} in both monocular~\cite{wang2024simtoreal} and panorama~\cite{wang2024lookahead} settings. \textbf{R2R-CE}
is collected based on the 
Matterport3D~\cite{chang2017matterport3d} scenes
with the Habitat simulator~\cite{savva2019habitat}. The R2R-CE dataset includes 5,611 trajectories divided into 
train, validation seen, validation unseen, and test unseen splits. Each trajectory has three English instructions 
with an average path length of 9.89 meters and an average instruction length of 32 words. 
Several standard metrics~\cite{anderson2018vision} \gh{are used} to evaluate VLN performance: 
Navigation Error (\textbf{NE}), Success Rate
(\textbf{SR}), SR given the Oracle stop policy (\textbf{OSR}), Success Rate weighted by normalized inverse Path Length (\textbf{SPL}).

\vspace{2mm}
\noindent \textbf{Zero-shot Object Navigation.} For object navigation, we evaluate our approach using the Habitat 
simulator~\cite{savva2019habitat} on the validation splits of two different datasets HM3D~\cite{ramakrishnan2habitat} and MP3D~\cite{chang2017matterport3d}. The \textbf{HM3D} validation split contains
2,000 episodes across 20 scenes and 6 object categories. The \textbf{MP3D} validation split contains 2,195 episodes across 11 scenes and 21 object categories. The main metrics~\cite{anderson2018vision} include Success Rate
(\textbf{SR}) and Success Rate weighted by normalized inverse Path Length (\textbf{SPL}).

\vspace{2mm}
\noindent \textbf{Situated Question Answering.} \gh{Following ScanNet~\cite{dai2017scannet},} the \textbf{SQA3D} dataset
comprises 20.4k descriptions and 33.4k diverse questions, which is splited into
train, val, and test sets. 
The main metric is the \textit{Exact Match} (\textbf{EM@1}) of the answer. Additionally, for localization evaluation, \textbf{Acc@0.5m} and \textbf{Acc@1.0m}  metric means 
\gh{the prediction is counted as correct when the predicted position is within 0.5 meter and 1.0 meter range to the ground truth position. }
The \textbf{Acc@15°} and \textbf{Acc@30°} metric means 
\gh{the prediction is counted as correct when the prediction orientation is within 15° and 30° range to the ground truth orientation.}

\subsection{Comparison with SOTA Methods}
As shown in Table~\ref{vln_monocular} and Table~\ref{vln_panorama}, we evaluate the VLN performance of 
\gh{our \OurModel{}} model on the R2R-CE dataset in both monocular and panorama settings, respectively. 
\gh{Table~\ref{vln_monocular} shows that our \OurModel{}} significantly outperforms previous monocular VLN methods on the Success Rate (SR) metric, even compared to LLM-based methods such as NaVid~\cite{zhang2024navid} and InstructNav~\cite{long2024instructnav}. Compared to the panorama setting, monocular VLN 
\gh{has the advantage of being} compatible with a broader range of real-world monocular robots. 
\gh{Our \OurModel{}} model overcomes the limitations of monocular cameras, enhancing the multi-view and BEV perception capabilities of the agent for monocular VLN.

\begin{table}[ht]
\vspace{-5pt}
\small
\tabcolsep=0.03cm
\centering{}%
\begin{tabular}{c|c|cccc|cccc}
\hline 
\multirow{2}{*}{Methods} & \multirow{2}{*}{LLM}  & \multicolumn{4}{c|}{Val Unseen} & \multicolumn{4}{c}{Test Unseen}\tabularnewline
 \cline{3-10} \cline{4-10} \cline{5-10} \cline{6-10} \cline{7-10} \cline{8-10} \cline{9-10} \cline{10-10} 
  & & NE\textdownarrow{} & OSR\textuparrow{} & SR\textuparrow{} & SPL\textuparrow{} & NE\textdownarrow{} & OSR\textuparrow{} & SR\textuparrow{} & SPL\textuparrow{}\tabularnewline
\hline

CM$^{2}$~\cite{georgakis2022cross} & $\times$ & 7.02 & 41.5 & 34.3 & 27.6  & 7.7 & 39 & 31 & 24\tabularnewline

WS-MGMap~\cite{chen2022weakly} & $\times$ & 6.28 & 47.6 & 38.9 & 34.3 & 7.11 & 45 & 35 & 28\tabularnewline

NaVid~\cite{zhang2024navid} & \ding{51} & \textbf{5.47} & 49.1 & 37.4 & \textbf{35.9} &  - & - & - & - \tabularnewline

InstructNav$^{*}$~\cite{long2024instructnav} & \ding{51} & 6.89 & - & 31 & 24 &  - & - & - & - \tabularnewline
 
\textcolor{black} VLN-3DFF~\cite{wang2024simtoreal} & $\times$ & 5.95 & 55.8 & 44.9 & 30.4 & 6.24 & 54.4 & 43.7 & 28.9
\tabularnewline
\hline

\textcolor{black} 
\OurModel{} (Ours) & $\times$ & 5.70 & \textbf{59.5} & \textbf{47.2} & 34.6 & \textbf{6.00} & \textbf{57.5} & \textbf{46.3} & \textbf{32.2}
\tabularnewline
\hline 
\end{tabular}
\vspace{-3pt}
\caption{Evaluation of VLN on 
R2R-CE 
with \textbf{monocular} setting. $*$ denotes zero-shot method.}\label{vln_monocular}
\vspace{-5mm}
\end{table}

\gh{We follow HNR~\cite{wang2024lookahead} to perform lookahead exploration through predicted candidate waypoint representations for the panorama setting in Table~\ref{vln_panorama}. Although the results show minor performance gains and the advanatges are not as pronounced as its monocular counterpart in Table~\ref{vln_monocular},}
\gh{our \OurModel{}} model \gh{still achieves} 
SOTA performance on the SPL metric and demonstrated competitive results on the SR metric. 

\begin{table}[ht]
\vspace{-5pt}
\small
\tabcolsep=0.03cm
\centering{}%
\begin{tabular}{c|c|cccc|cccc}
\hline 
\multirow{2}{*}{Methods} & \multirow{2}{*}{LLM}  & \multicolumn{4}{c|}{Val Unseen} & \multicolumn{4}{c}{Test Unseen}\tabularnewline
 \cline{3-10} \cline{4-10} \cline{5-10} \cline{6-10} \cline{7-10} \cline{8-10} \cline{9-10} \cline{10-10} 
  & & NE\textdownarrow{} & OSR\textuparrow{} & SR\textuparrow{} & SPL\textuparrow{} & NE\textdownarrow{} & OSR\textuparrow{} & SR\textuparrow{} & SPL\textuparrow{}\tabularnewline
\hline

Sim2Sim~\cite{krantz2022sim2sim} & $\times$ & 6.07 &  52 & 43 & 36  & 6.17 & 52 & 44 & 37\tabularnewline

VLN-BERT~\cite{Hong2022bridging} & $\times$ & 5.74 & 53 & 44 & 39  & 5.89 & 51 & 42 & 36\tabularnewline

GridMM~\cite{wang2023gridmm} & $\times$ & 5.11 & 61 & 49 & 41 & 5.64 & 56 & 46 & 39\tabularnewline

Ego$^{2}$-Map\textcolor{black}~\cite{hong2023learning} & $\times$ & 4.94 & - & 52 & 46 & 5.54 & 56 & 47 & 41 \tabularnewline

DREAM~\cite{wang2023dreamwalker} & $\times$ & 5.53 & 59 & 49 & 44 & 5.48 & 57 & 49 & 44\tabularnewline

ScaleVLN\textcolor{black}\ \cite{wang2023scaling} & $\times$ & 4.80 & - & 55 & 51 & 5.11 & - & 55 & 50 \tabularnewline

ETPNav~\cite{an2024etpnav} & $\times$ & 4.71 & 65 & 57 & 49 &  5.12 & 63 & 55 & 48 \tabularnewline

BEVBert~\cite{an2023bevbert} & $\times$ & 4.57 & 67 & 59 & 50 & \textbf{4.70} & 67 & \textbf{59} & 50 \tabularnewline

\textcolor{black} HNR~\cite{wang2024lookahead} & $\times$ & \textbf{4.42} & 67 & \textbf{61} & 51 & 4.81 & 67 & 58 & 50
\tabularnewline

\textcolor{black} Energy~\cite{liu2024vision} & $\times$ & 4.69 & 65 & 58 & 50 & 5.08 & 64 & 56 & 48
\tabularnewline
\hline

\textcolor{black} 
\OurModel{} (Ours) & $\times$ & 4.53 & \textbf{68} & \textbf{61} & \textbf{52} & 4.78 & \textbf{68} & 58 & \textbf{51}
\tabularnewline
\hline 
\end{tabular}
\vspace{-3pt}
\caption{Evaluation of VLN on 
R2R-CE 
with \textbf{panorama} setting.}\label{vln_panorama} 
\vspace{-1mm}
\end{table}

In Table~\ref{obj_navigation} for the Zero-shot Object Navigation, 
\gh{our \OurModel{}} achieves SOTA performance in the SPL metric and 
achieves competitive results in the SR metric. Notably, 
\gh{our \OurModel{}} is the only method that queries targets using 
feature fields instead of VLM. Replacement of BLIP-2~\cite{li2023blip} in VLFM~\cite{yokoyama2024vlfm} with 
\OurModel{} improves the navigation success rate (SR) by nearly 3\%. Although the MP3D benchmark includes some targets outside the \OurModel{} object vocabulary, our model still performs well, demonstrating strong generalization. 
Compared to methods using LLM: 
InstructNav~\cite{long2024instructnav} and SG-Nav~\cite{yin2024sgnavonline3dscene}, 
\gh{our \OurModel{} also} offers significant advantages in response time and computational cost. 

\begin{table}[ht]
\small
\tabcolsep=0.04cm
\centering{}%
\begin{tabular}{c|c|c|c|cc|cc}
\hline 
\multirow{2}{*}{Methods} & \multirow{2}{*}{LLM} & \multirow{2}{*}{VLM}  & \multirow{2}{*}{Feature Fields} & \multicolumn{2}{c|}{HM3D} & \multicolumn{2}{c}{MP3D}\tabularnewline
 \cline{5-8} \cline{6-8} \cline{7-8} \cline{8-8} 
  & & & &  SR\textuparrow{} & SPL\textuparrow{} &  SR\textuparrow{} & SPL\textuparrow{}
\tabularnewline
\hline

\textcolor{black} ZSON~\cite{majumdar2022zson} & $\times$ & \ding{51} & $\times$ & 25.5 & 12.6 & 15.3 & 4.8 
\tabularnewline

\textcolor{black} ESC~\cite{zhou2023esc} & \ding{51} & \ding{51} & $\times$ & 39.2 & 22.3 &  28.7 & 14.2 
\tabularnewline

\textcolor{black} VLFM~\cite{yokoyama2024vlfm} & $\times$ & \ding{51} & $\times$ & 52.5 & 30.4 &  36.4 & 17.5 
\tabularnewline

\textcolor{black} InstructNav~\cite{long2024instructnav} & \ding{51} & \ding{51} & $\times$ & \textbf{58.0} & 20.9 &  - & - 
\tabularnewline

\textcolor{black} GAMap~\cite{yuan2024gamapzeroshotobjectgoal} & \ding{51} & \ding{51} & $\times$ & 53.1 & 26.0 &  - & - 
\tabularnewline

\textcolor{black} SG-Nav~\cite{yin2024sgnavonline3dscene} & \ding{51} & \ding{51} & $\times$ & 54.0 & 24.9 & \textbf{40.2} & 16.0 
\tabularnewline

\hline

\textcolor{black} 
\OurModel{} (Ours) & $\times$ & $\times$ & \ding{51} & 55.6 &  \textbf{31.8} & 39.0 & \textbf{18.8}
\tabularnewline
\hline

\hline 
\end{tabular}
\vspace{-5pt}
\caption{Evaluation of Zero-shot Object Navigation on the HM3D and MP3D benchmarks.}\label{obj_navigation}
\end{table}
\vspace{-1mm}

In Table~\ref{sqa3d} for the Situated Question Answering task, 
\gh{our \OurModel{}} achieves good localization performance in metrics 
\gh{of} Acc@0.5m, Acc@1m, Acc@15° and Acc@30°. Although our performance on the answering accuracy (EM@1) is significantly lower than that of LLM-based methods: 
LEO~\cite{huang2023embodied} and Scene-LLM~\cite{fu2024scene}, it is worth noting that 
\gh{our \OurModel{}} \textit{\textbf{only uses images}} as input without low-noise 3D point clouds. This actually offers a significant advantage in agent-centered embodied tasks since it is more adaptable to unseen dynamic real-world environments, where the low-noise point clouds are difficult to collect.

\begin{table}[ht]
\vspace{-5pt}
\small
\tabcolsep=0.04cm
\centering{}%
\begin{tabular}{c|c|c|c|cc|cc|c}
\hline 
\multirow{2}{*}{Methods} & \multirow{2}{*}{LLM} & \multirow{2}{*}{PCD}  & \multirow{2}{*}{Image} & \multicolumn{2}{c|}{Position} & \multicolumn{2}{c|}{Orientation} & \multicolumn{1}{c}{Answer}\tabularnewline
 \cline{5-9} \cline{6-9} \cline{7-9} \cline{8-9} \cline{9-9} 
  & & & &  0.5m  &  1.0m  &  15° & 30° & EM@1
\tabularnewline
\hline

\textcolor{black} ClipBERT~\cite{lei2021less} & $\times$ & $\times$ & \ding{51} & - &  - & - &  - & 43.3
\tabularnewline

\textcolor{black} ScanQA~\cite{azuma2022scanqa} & $\times$ & \ding{51} & $\times$ & - &  - & - &  - & 46.6
\tabularnewline

\textcolor{black} SQA3D~\cite{masqa3d} & $\times$ & \ding{51} & $\times$ & 14.6 &  34.2 & 22.4 &  42.3 & 47.2
\tabularnewline

\textcolor{black} 3D-VisTA~\cite{zhu20233d} & $\times$ & \ding{51} & $\times$ & - &  - & - &  - & 48.5
\tabularnewline

\textcolor{black} SceneVerse~\cite{jia2024sceneverse} & $\times$ & \ding{51} & $\times$ & - &  - & - &  - & 49.9
\tabularnewline

\textcolor{black} LEO~\cite{huang2023embodied} & \ding{51} & \ding{51} & $\times$ & - &  - & - &  - & 52.4
\tabularnewline

\textcolor{black} Scene-LLM~\cite{fu2024scene} & \ding{51} & \ding{51} & \ding{51} & - &  - & - &  - & \textbf{54.2}
\tabularnewline

\hline

\textcolor{black} 
\OurModel{} (Ours) & $\times$ & $\times$ & \ding{51} & \textbf{23.4}
 &  \textbf{45.7} & \textbf{29.8}
 & \textbf{54.7} & 47.7

\tabularnewline
\hline

\hline 
\end{tabular}
\vspace{-3pt}
\caption{Evaluation of Situated Question Answering (SQA3D) task. \textbf{PCD} denotes methods that use point clouds as input, while \textbf{Image} represents methods that use images as input. }
\label{sqa3d}
\vspace{-4mm}
\end{table}

\subsection{Ablation Study}
\noindent \textbf{
\gh{Perfromance impact of \OurModel{} on embodied tasks.}
} 
In row 1 of Table~\ref{ablation_modules}, 
the performance of monocular VLN and object navigation drops significantly \gh{without representations from \OurModel{}}. In this setting, the VLN model only uses the CLIP features from the forward-facing view with 
\gh{features of all other directions} set to zero. The object navigation model uses BLIP-2~\cite{li2023blip} instead of \OurModel{} to construct the value map. Examining rows 2 and 3 shows that removing either the novel view or the BEV map reduces the performance of both two tasks, highlighting the role of each \OurModel{} module.

\vspace{2mm}
\noindent \textbf{Novel views are crucial for monocular VLN.}
As shown in row 1 and row 2 of Table~\ref{ablation_modules},  the novel view representations significantly boost VLN performance by overcoming the narrow perception of the monocular camera~\cite{wang2024simtoreal}, enabling the monocular agent to have panoramic perception capabilities. 
\gh{To some extent, this} confirms that novel view prediction is a very important and valuable capability for monocular agents. Based on this capability, the \OurModel{} model predicts the novel view representations of candidate waypoints around the agent to construct the topological map for better navigation planning.

\vspace{2mm}
\noindent \textbf{Object navigation requires balancing local and global targets.} As shown in row 3 of Table~\ref{ablation_modules}, we observe that relying solely on BEV representation significantly reduces object navigation performance. This decline occurs because the global value map from the BEV map fails to select optimal nearby waypoints if the target is far from these waypoints. In this case, a local value map constructed from novel views is also essential to identify the optimal short-term goal, \textit{i.e.}, nearby waypoints around the agent.

\begin{table}[ht]
\vspace{-5pt}
\small
\tabcolsep=0.04cm
\centering{}%
\begin{tabular}{c|c|cccc|cc}
\hline 
\multirow{2}{*}{View \& Pano}  & \multirow{2}{*}{BEV}  &\multicolumn{4}{c|}{Monocular VLN} & \multicolumn{2}{c}{Object Nav.}\tabularnewline
\cline{3-8} \cline{4-8} \cline{5-8} \cline{6-8} \cline{7-8} \cline{8-8} 
  & & NE\textdownarrow{} & OSR\textuparrow{} & SR\textuparrow{} & SPL\textuparrow{} & SR\textuparrow{} & SPL\textuparrow{}\tabularnewline
\hline 

$\times$ & $\times$ & 6.54 & 44.6 & 33.1 & 23.4
 & 52.5 & 30.4
\tabularnewline

\ding{51} & $\times$ & 5.78 & 58.3 & 46.9 & 32.7 & 53.9 & 30.8
\tabularnewline

$\times$ & \ding{51} & 6.02 & 53.1 & 42.8 & 26.5 & 50.2 & 27.1
\tabularnewline

\ding{51} & \ding{51} & \textbf{5.70} & \textbf{59.5} & \textbf{47.2} & \textbf{34.6} & \textbf{55.6} & \textbf{31.8}

\tabularnewline
\hline 
\end{tabular}
\vspace{-5pt}
\caption{Ablation study for the modules of \OurModel{}.}\label{ablation_modules}
\vspace{-1mm}
\end{table}

\begin{table}[ht]
\small
\tabcolsep=0.04cm
\centering{}%
\begin{tabular}{c|c|c|cccc|cc}
\hline 
\multirow{2}{*}{OBJ-CL} & \multirow{2}{*}{CLIP-CL} & \multirow{2}{*}{FG-CL}  &\multicolumn{4}{c|}{Monocular VLN} & \multicolumn{2}{c}{Object Nav.}\tabularnewline
\cline{4-9} \cline{5-9} \cline{6-9} \cline{7-9} \cline{8-9} \cline{9-9} 
  & & & NE\textdownarrow{} & OSR\textuparrow{} & SR\textuparrow{} & SPL\textuparrow{} & SR\textuparrow{} & SPL\textuparrow{}\tabularnewline
\hline 

$\times$ & $\times$ & $\times$ & 6.21 & 50.2 & 40.7 & 24.9 & 34.2 & 13.9
\tabularnewline

$\times$ & \ding{51} & $\times$ & 5.84 & 56.1 & 44.6 & 31.1 & 47.6 & 27.8
\tabularnewline

\ding{51} & $\times$ & \ding{51} & 6.01 & 53.5 & 42.4 & 26.7 & \textbf{55.8} & 31.6
\tabularnewline

unbalanced & \ding{51} & \ding{51} & 5.73 & 58.3 & 46.6 & 33.0 & 51.7 & 28.8
\tabularnewline

\ding{51} & \ding{51} & coarse & 5.81 & 57.1 & 45.7 & 33.2 & 55.5 & 31.2
\tabularnewline

\ding{51} & \ding{51} & \ding{51} & \textbf{5.70} & \textbf{59.5} & \textbf{47.2} & \textbf{34.6} & 55.6 & \textbf{31.8}

\tabularnewline
\hline 
\end{tabular}
\vspace{-5pt}

\caption{Ablation study for the multi-level contrastive pre-training. \textbf{OBJ-CL}: object-level contrastive learning. \textbf{CLIP-CL}: knowledge distillation using CLIP visual features from ground-truth view. \textbf{FG-CL}: fine-grained contrastive learning for long text understanding.}\label{ablation_CL}
\vspace{-1mm}
\end{table}

\vspace{2mm}
\noindent \textbf{Pre-training is essential for generalizable feature fields model.}
Table~\ref{ablation_CL} analyzes the impact of multi-level contrastive pre-training on downstream embodied tasks. As shown in row 1 of Table~\ref{ablation_CL}, 
\gh{the performance on VLN and object navigation drops significantly when the} model is optimized solely by the navigation loss~\cite{an2024etpnav} \gh{without pre-training.} 

\vspace{2mm}
\noindent \textbf{Both CLIP distillation and language supervision are indispensable.} For row 3 of Table~\ref{ablation_CL} without supervision from the CLIP visual features, the VLN performance lags behind 
the model distilled by CLIP. This suggests that millions of language annotations are still far from sufficient for \OurModel{} pre-training, and distilling representations from 2D foundation models to enhance semantic generalization remains necessary. However, in Table~\ref{ablation_CL}, we can also see that language supervision significantly improves \OurModel{} performance on embodied tasks 
\gh{, the model performs poorly in row 2 when using only CLIP distillation.}

\vspace{2mm}
\noindent \textbf{Long-tail distribution limits object-level semantic learning.} As shown in row 4 of Table~\ref{ablation_CL}, 
the performance of object navigation decreases 
\gh{drastically without the balanced loss mentioned in Section~\ref{sec:multi_CL}}. The long-tail distribution of object categories in indoor environments leads models to overlook \gh{of} rare or small objects 
\gh{such as} \textit{towels} and \textit{cups}, significantly limiting 
the ability \gh{of our \OurModel{} model} to query target objects. Fortunately, row 6 of Table~\ref{ablation_CL} shows that the balanced object alignment works well by balancing the weight for loss of hard-to-recognize objects.

\vspace{2mm}
\noindent \textbf{Fine-grained contrastive benefits long text understanding.} In the row 5 of Table~\ref{ablation_CL}, we use the [SEP] feature (single vector) from the CLIP text encoder to supervise panorama and BEV representations. However, compared to the fine-grained contrastive learning in row 6, compressing long text into a coarse vector significantly limits \OurModel{}'s performance on long-text understanding tasks such as VLN. As shown in Figure~\ref{fig:framework}, fine-grained contrastive learning between long texts and windows within the BEV map helps \OurModel{} understand spatial layouts, overcoming the limitations of semantic representation in large-scale scenes.

\begin{table}[ht]
\vspace{-3pt}
\small
\tabcolsep=0.04cm
\centering{}%
\begin{tabular}{c|c|c|c|c}
\hline 
\multirow{1}{*}{Rays for View} &\multirow{1}{*}{View} & \multirow{1}{*}{Panorama} & \multirow{1}{*}{Rays for BEV} & \multirow{1}{*}{BEV} \tabularnewline
\hline 

73.6 FPS & 71.1 FPS & 5.9 FPS  & 6.3 FPS & 6.1 FPS

\tabularnewline
\hline 
\end{tabular}
\vspace{-3pt}
\caption{Runtime analysis measured on one RTX 4090 GPU. \textbf{FPS} denotes Frames Per Second.}\label{runtime}
\vspace{-1mm}
\end{table}

\noindent \textbf{
\OurModel{} enables real-time inference.} As shown in Table~\ref{runtime}, we calculate the inference time of 
\gh{our \OurModel{}} model on the val unseen split of the R2R-CE dataset in the VLN task. 
\gh{Our \OurModel{}} achieves novel view volume rendering at 73.6 FPS, which slightly drops to 71.1 FPS when rays are further encoded by the View Encoder. For a panorama containing 12 views, the inference speed is 5.9 FPS. Due to the large rendered range, 
our \OurModel{} renders BEV maps at 6.3 FPS, which drops slightly to 6.1 FPS with the BEV Map Encoder. 
Our \OurModel{} model adopts the same \textit{sparse sampling} strategy as in HNR~\cite{wang2024lookahead}, where the MLP network is only used to render sampled regions containing feature points nearby, while skipping empty regions. This reduces rendering time by over 10 times, enabling real-time embodied tasks.
\section{Conclusion}
\label{sec:conclusion}

In this work, we propose Generalizable 3D-Language Feature Fields (
\OurModel{}), a 3D representation model pre-trained on large-scale 3D-language data for embodied tasks. We organize the first large-scale 3D-language dataset for feature fields training, demonstrating the feasibility of using generalizable feature fields for large-scale scene understanding, \ie, panorama and BEV. \gh{Our proposed \OurModel{} leverages} 
multi-level contrastive learning strategies such as balanced object semantic alignment, fine-grained text alignment, and CLIP knowledge distillation 
\gh{to optimize} generalized feature fields. More importantly, the value of \OurModel{} has been widely evaluated in multiple embodied tasks. We believe that 
our \OurModel{} can provide sufficient inspiration for subsequent research on feature fields and embodied AI.

\vspace{2mm}
\noindent \textbf{Limitations and future works.} 
\gh{Our \OurModel{}} still has some limitations 
\gh{with} significant potential for future research: 1) 
\OurModel{} cannot be adapted to dynamic environments, where objects or people are moving in real time. 
\gh{This} requires better update strategies for implicit representations. 2) 
\OurModel{} has not been evaluated on dynamic tasks such as object manipulation. 3) The scale and quality of 3D-language data used for training 
\OurModel{} remain limited, which essentially restricts the ability of generalizable feature field models. 4) The 3D feature fields combined with LLM 
can enable better text generation. These may become the guiding directions for the next phase of generalizable feature fields.

\clearpage

{\small
\bibliographystyle{ieeenat_fullname}
\bibliography{11_references}
}

\clearpage \appendix 


\maketitlesupplementary

\section{More Details of the \OurModel{} Model}
\label{sec:details_of_the_model_structure}

\begin{figure}
\noindent\begin{minipage}[h]{1\columnwidth}%
\begin{center}
\includegraphics[width=1.0\columnwidth]{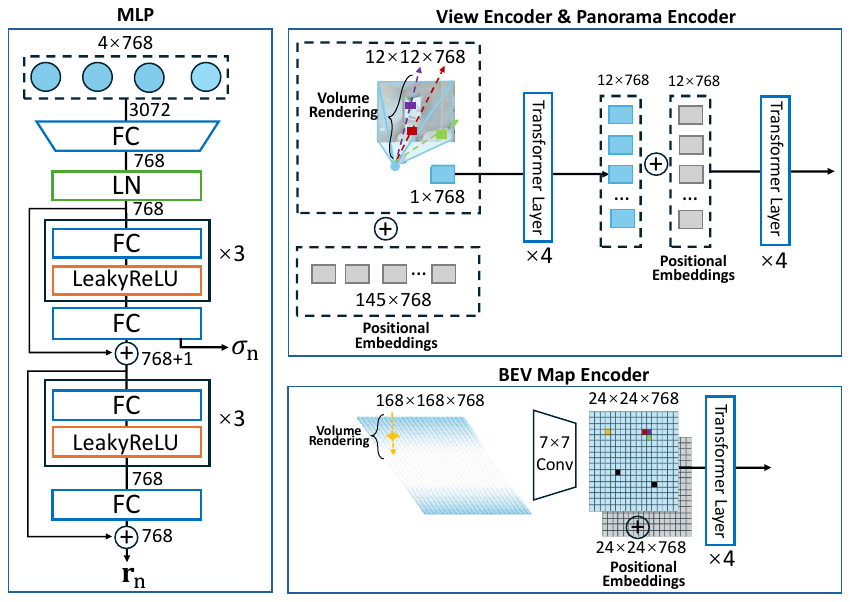}
\par\end{center}%
\end{minipage}
\vspace{-3mm}
\caption{Architecture of modules in the \OurModel{} model. FC denotes
a fully connected layer, LN denotes layer normalization and LeakyReLU~\cite{maas2013rectifier} is the activation function.}
\label{fig:structure}
\vspace{-5pt}
\end{figure}

\paragraph{Model structure.} Figure~\ref{fig:structure} illustrates the structure of main modules in the \OurModel{} model. Compared to HNR~\cite{wang2024lookahead}, \OurModel{} improve the MLP network for volume rendering by adding residual connections and replacing ReLU with LeakyReLU, which helps alleviate gradient explosion and neuron death issues during HNR training. Since the number of k-nearest features is set to 4 and the dimension of each aggregated feature is 768, the input dimension of both $\textbf{MLP}_{view}$ and $\textbf{MLP}_{BEV}$ networks is 3072. As shown in Figure~\ref{fig:structure}, all transformer-based encoders consist of four-layer transformers.

\vspace{-15pt}
\paragraph{Settings of novel view prediction.} For each sampled point in the rendered ray, we set the search radius for k-nearest features as 0.5 meter. Using \textit{sparse sampling}~\cite{wang2024lookahead}, if no nearby feature points are found within a sampled point's search radius, the latent feature and volume density are set to zero. The rendered ray is uniformly sampled from 0 to 10 meters, and the number of sampled points is set as 501. After volume rendering, the number of rays within a novel view is set as 12$\times$12. 

\vspace{-15pt}
\paragraph{Settings of BEV map prediction.} The search radius for k-nearest features is set as 0.4 meter. The rendered ray is uniformly sampled from 0 to 1.6 meters (\textit{i.e.}, vertically from the camera's position to bottom), and the number of sampled points is set as 17. After volume rendering, the number of rays within a BEV map is set as 168$\times$168.

\begin{figure}
\noindent\begin{minipage}[hb]{1\columnwidth}%
\begin{center}
\includegraphics[width=1.0\columnwidth]{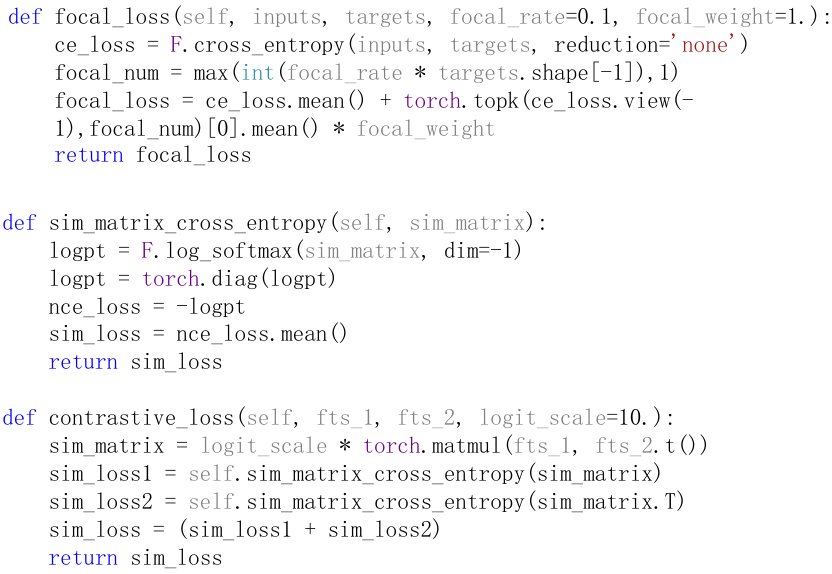}
\par\end{center}%
\end{minipage}
\vspace{-5pt}
\caption{PyTorch implementation of loss functions for the balanced object semantic alignment and the CLIP knowledge distillation.}
\label{fig:loss_1}
\vspace{-5pt}
\end{figure}

\begin{figure*}[hb]
\makebox[\textwidth][c]
{\includegraphics[width=0.8\paperwidth]{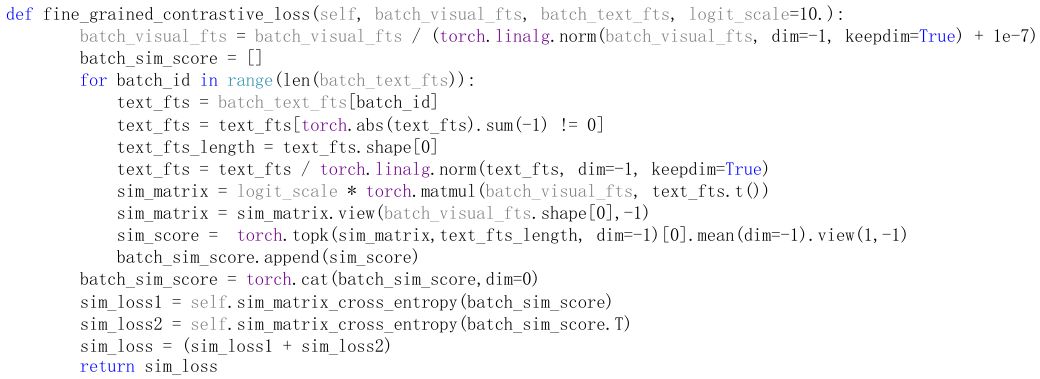}}
\vspace{-15pt}
\caption{PyTorch implementation of loss function for the fine-grained contrastive learning.}
\label{fig:loss_2}
\vspace{-5pt}
\end{figure*}

\vspace{-15pt}
\paragraph{Loss functions.} As illustrated in Figure~\ref{fig:loss_1} and \ref{fig:loss_2}, we present the code for the primary loss functions used in \OurModel{} pre-training to provide further details. During training, we apply constant coefficients to balance the contributions of each loss, ensuring they remain within the same order of magnitude.

\section{Visualization of the Training Data}
\begin{figure*}[hb]
\makebox[\textwidth][c]
{\includegraphics[width=0.8\paperwidth]{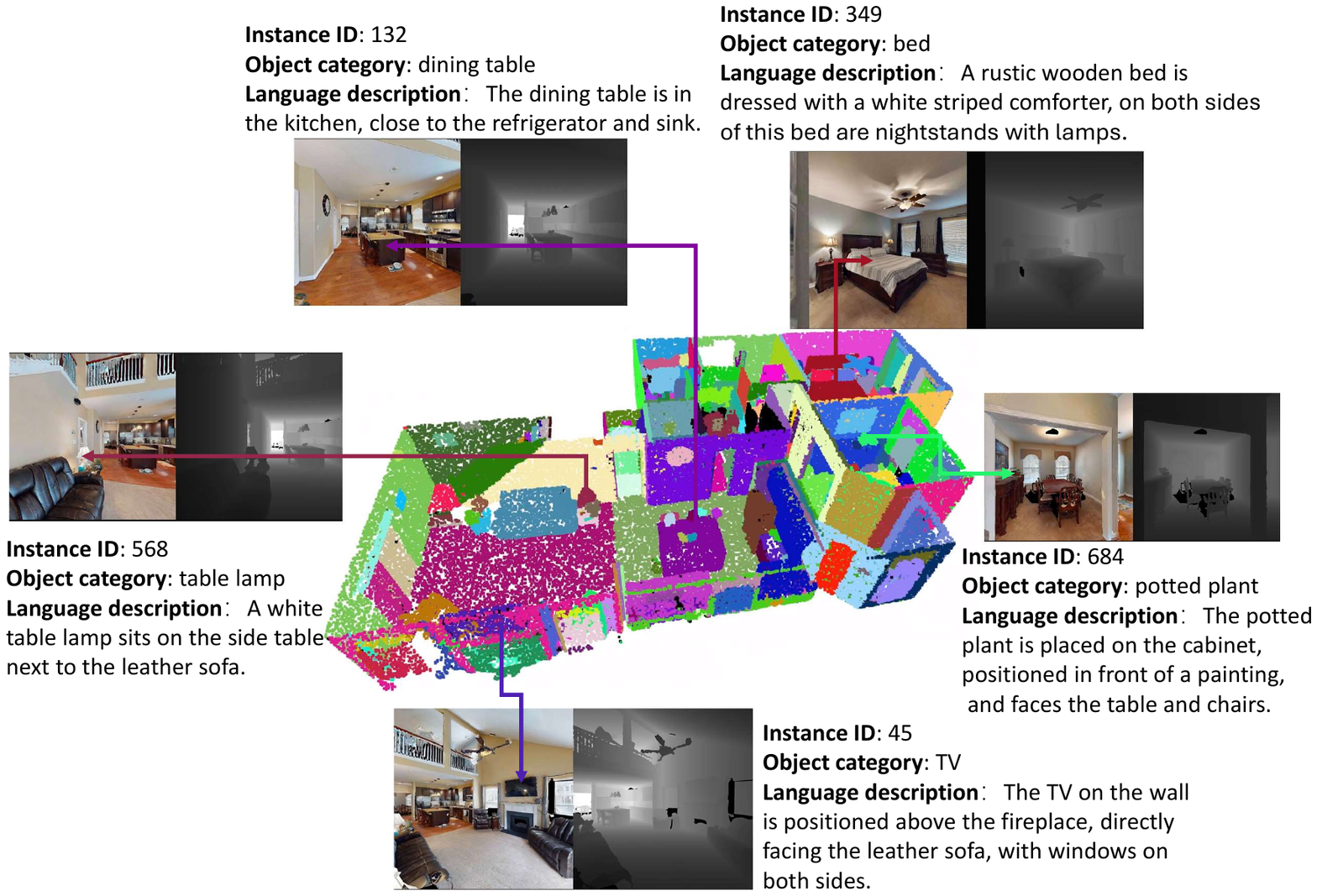}}
\vspace{-20pt}
\caption{Demonstration of a 3D scene in the training data. Instance-level point clouds mark all instances with object categories, and some instances enriched with language descriptions.}
\label{fig:dataset}
\vspace{-10pt}
\end{figure*}

As shown in Figure~\ref{fig:dataset}, we present a 3D scene from our dataset along with some associated language annotations (scene 00800-TEEsavR23oF from HM3D~\cite{ramakrishnan2habitat}). The instance-level point cloud precisely annotates instances within the 3D scene, allowing retrieval of language annotations for any position by calculating its neighboring instance points and using the instance IDs.

\section{Visualization of the \OurModel{} model}
\begin{figure*}[ht]
\makebox[\textwidth][c]
{\includegraphics[width=0.8\paperwidth]{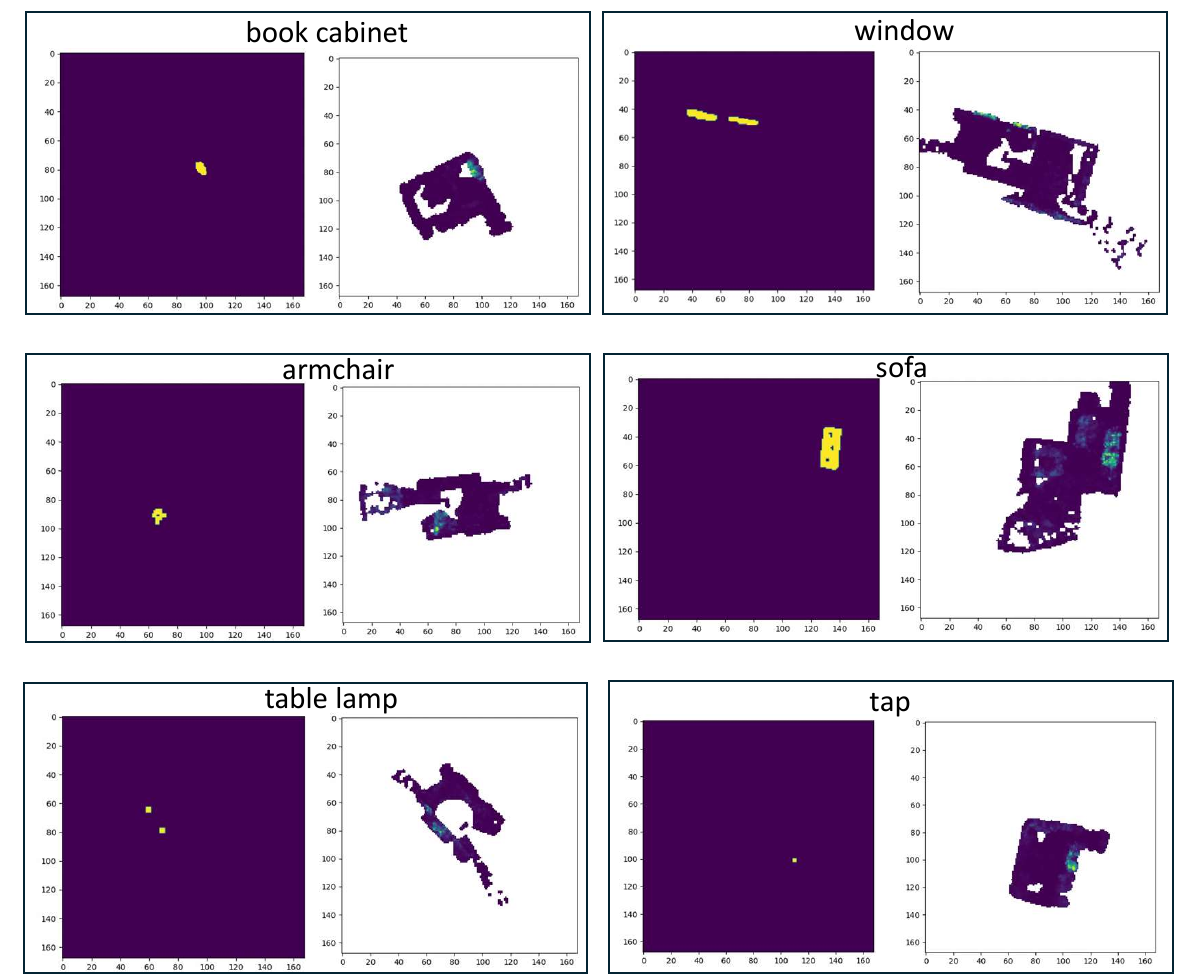}}
\vspace{-20pt}
\caption{Visualization of querying objects on rays of the \OurModel{}'s BEV map. The left side of each example is GT, and the right side is the query result. Please zoom in for a better view.}
\label{fig:object_visualization}
\vspace{-10pt}
\end{figure*}

\begin{figure*}[ht]
\makebox[\textwidth][c]
{\includegraphics[width=0.8\paperwidth]{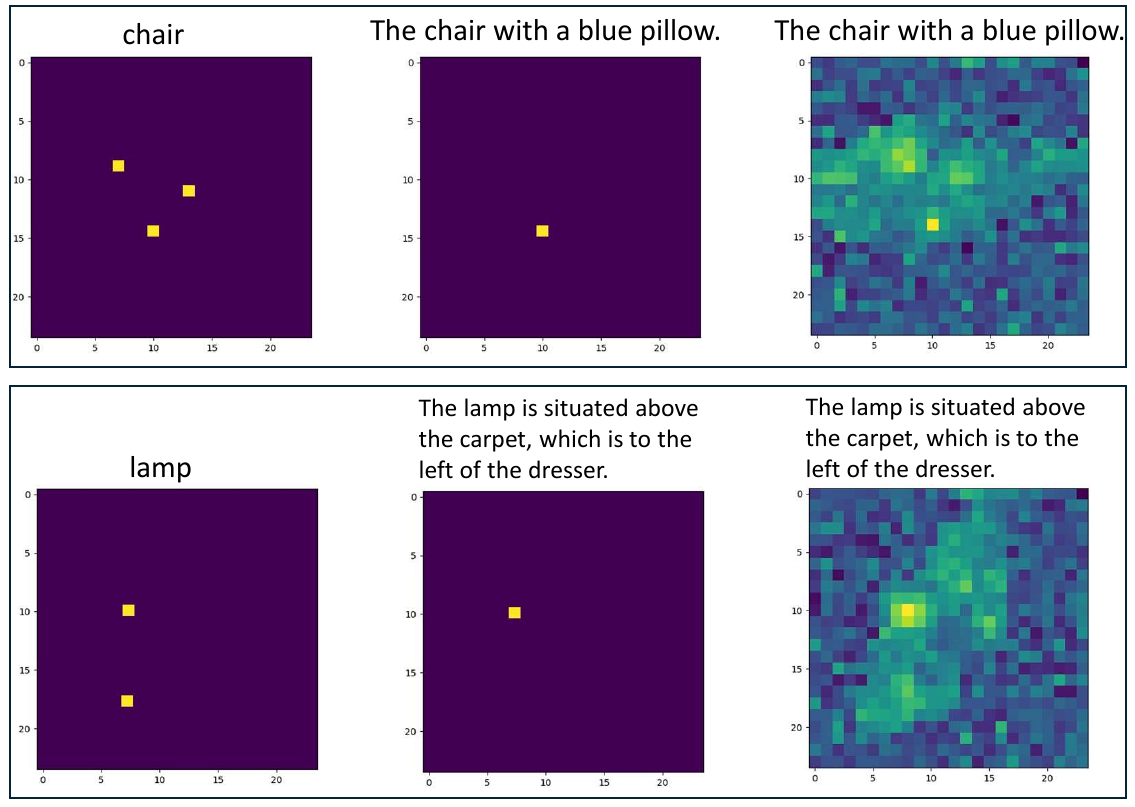}}
\vspace{-20pt}
\caption{Visualization of querying long texts on the BEV map of our \OurModel{}. Each example has the object’s GT on the left, the long text GT in the middle, and the query result of the long text on the right. Please zoom in for a better view.}
\label{fig:text_visualization}
\vspace{-10pt}
\end{figure*}

As shown in Figure~\ref{fig:object_visualization} and \ref{fig:text_visualization}, the \OurModel{} model query targets with language on the BEV map. In Figure ~\ref{fig:object_visualization}, the left side of each example shows the position of the ground-truth target, while the right side displays the result of querying objects on rays of the BEV map during navigation. The BEV map accurately recognizes both large objects, like \textit{window} and \textit{sofa}, and smaller objects, like \textit{table lamp} and \textit{tap}, by calculating the cosine similarity between ray representations and target text features.

In Figure~\ref{fig:text_visualization}, the left side of each example shows the position of the objects, the middle is the ground-truth position of the long text that contains the target object, while the right side displays the result of querying the long text on the BEV map during navigation. In the 3D scene, multiple objects of the same category often appear. With the excellent ability to understand long texts, our g3D-LF model can achieve more fine-grained long-text queries, distinguishing different instances of the same object category.

\end{document}